\documentclass{article} 
\usepackage{iclr2023_conference,times}

\usepackage{amsmath,amsfonts,bm}

\def\eqref#1{equation~\ref{#1}}

\def\1{\bm{1}}

\DeclareMathAlphabet{\mathsfit}{\encodingdefault}{\sfdefault}{m}{sl}
\SetMathAlphabet{\mathsfit}{bold}{\encodingdefault}{\sfdefault}{bx}{n}

\usepackage[utf8]{inputenc} 
\usepackage[T1]{fontenc}    
\usepackage{hyperref}       
\usepackage{url}            
\usepackage{booktabs}       
\usepackage{amsmath,amsfonts,amssymb}       
\usepackage{mathtools}
\usepackage{nicefrac}       
\usepackage{microtype}      
\usepackage{xcolor}         
\usepackage{bm}
\usepackage{dsfont}
\usepackage{algorithmic,algorithm}
\usepackage{graphicx}
\usepackage{cleveref}
\usepackage{wrapfig}

\usepackage{lipsum}
\usepackage{subcaption}

\title{Variation-based Cause Effect Identification}

\author{Mohamed Amine ben Salem  \\
Accenture GmbH \\
70178 Stuttgart, Germany \\
\texttt{mohamed.ben.salem@accenture.com} \\
\And
Karim Said Barsim \\
Robert Bosch GmbH \\
71272 Renningen, Germany \\
\texttt{karim.barsim@de.bosch.com} \\
\And
Bin Yang\\
University of Stuttgart \\
D-70569 Stuttgart, Germany \\
\texttt{bin.yang@iss.uni-stuttgart.de}
}

\iclrfinalcopy 
\begin{document}

\maketitle
\begin{abstract}
Mining genuine mechanisms underlying the complex data generation process in real-world systems is a fundamental step in promoting interpretability of (and thus trust in) data-driven models. 
Therefore, we propose a \underline{v}ariation-based \underline{c}ause \underline{e}ffect \underline{i}dentification (VCEI) framework for causal discovery in bivariate systems from a single observational setting.
Our framework relies on the principle of \underline{i}ndependence of \underline{c}ause and \underline{m}echanism (ICM) under the assumption of an existing acyclic causal link, and offers a practical realization of this principle.
Principally, we artificially construct two settings in which the marginal distributions of one covariate, claimed to be the cause, are guaranteed to have non-negligible variations.
This is achieved by re-weighting samples of the marginal so that the resultant distribution is notably distinct from this marginal according to some discrepancy measure.
In the causal direction, such variations are expected to have no impact on the effect generation mechanism.
Therefore, quantifying the impact of these variations on the conditionals reveals the genuine causal direction.
Moreover, we formulate our approach in the kernel-based maximum mean discrepancy, lifting all constraints on the data types of cause and effect covariates, and rendering such artificial interventions a convex optimization problem.
We provide a series of experiments on real and synthetic data showing that VCEI is, in principle, competitive to other cause effect identification frameworks.
\end{abstract}

\section{Introduction}
\label{sec:introduction}

Building trust in our machine learning models requires that they extend beyond their current limits of learning associational patterns and correlations.
We need to be able to use them in interacting with our surroundings, in taking action to change or improve our environment, or in querying them for hypothetical scenarios that requires transparency.
Yet, their black-box characteristics constitute significant barriers to their wide-scale adoption in, e.g., safety-critical domain.
Causal inference relies on genuine cause-effect relationships rather purely statistical associations, thus promoting our understanding of the underlying data generation process. 

While inferring genuine causal relations (oftentimes termed \emph{causal discovery}) is, in general, a challenging task, it is even more challenging in bivariate systems where many of the early methods (based on conditional independence tests \cite{spirtes2000causation,sun2007distinguishing,pearl2009causality}) fall short.
Moreover, bivariate causal discovery is a fundamental step in mining implicit asymmetries in larger structures.
In bivariate systems, asymmetries in the functional relationship (e.g. causal relationships tend to be functionally simpler, more elementary, and easier to learn with limited-capacity models than purely associational ones) is an example of a characteristic permitting identifiability of causal structure from observational data.

Another example of such an asymmetry is the postulate of independent mechanisms, on which our framework relies. 
In this principle, it is assumed that causal relationships tend to decompose into invariant, stable sub-mechanisms.
Such a principle has been the core asymmetry exploited in numerous bivariate causal discovery frameworks as shall be discussed in \cref{sec:related-work}.
In this work, we exploit a barely explored interpretation of this principle, namely that these sub-mechanisms do not influence each other. 
To this end, we introduce variations to cause generation mechanism and quantify the influence on the effect generation mechanism.

\begin{figure}[t]
	\begin{center}
		\includegraphics[width=\linewidth]{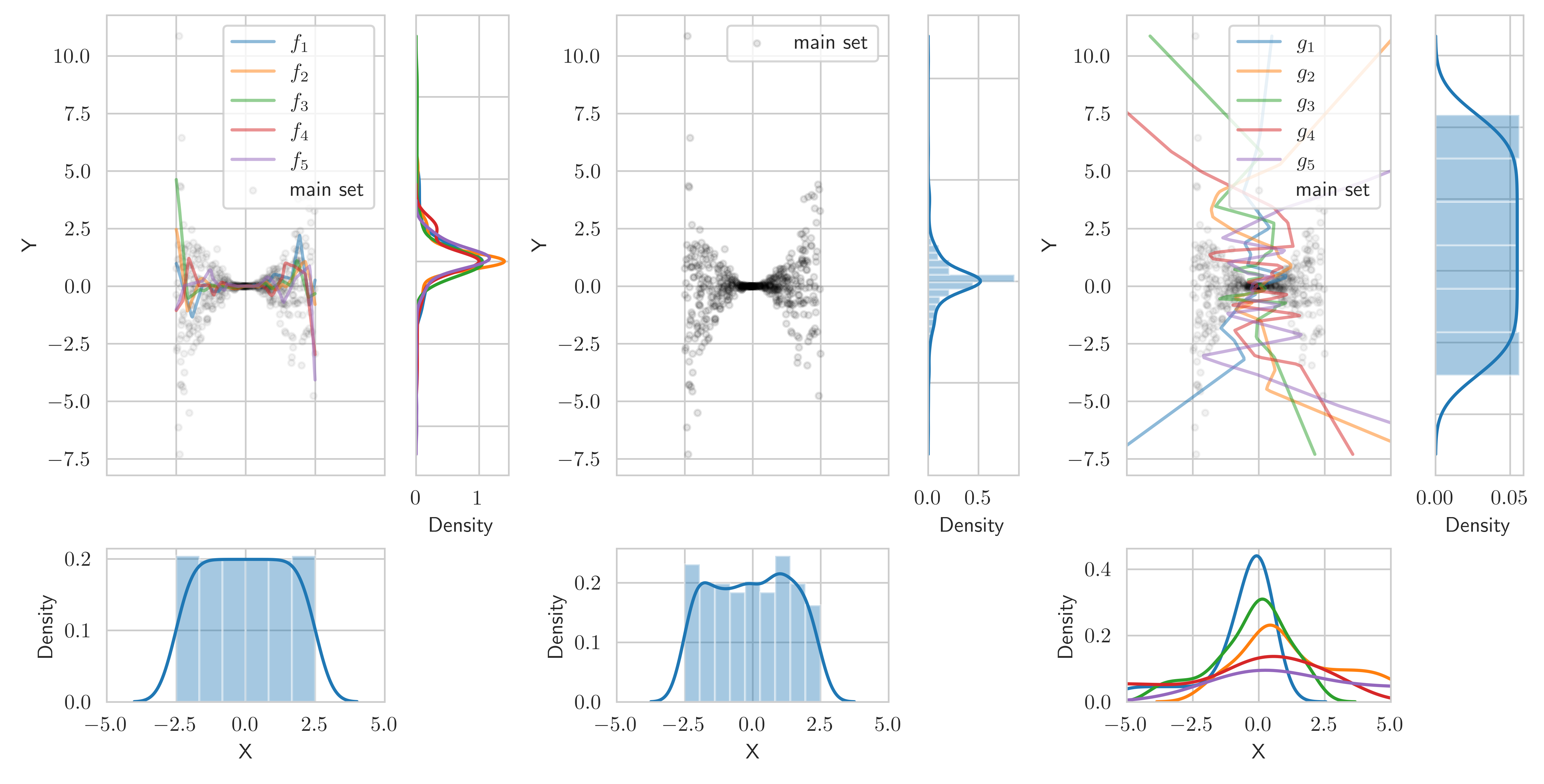}
		\caption{
a toy example illustrating the asymmetry induced by the principle of independent mechanisms, and the effect of variations.
The genuine data generation process is $y = -\frac{1}{2} x^2 * \epsilon$ with $x\sim\mathcal{U}[-2.5, 2.5]$ and the multiplicative noise $\epsilon$ from a standard normal distribution.
A sample set from such a process are depicted in the scatter plot (middle).
To na\"ivly introduce variations, we randomly draw a new set from the data generation process and train a model in the causal direction $f_i: x \mapsto f_i(x)$ (left), and similarly in the acausal direction $g_i: y: g_i(y)$ (right).
The figure illustrates stability of the causal predictive models compared to acausal ones.
}
\vspace{-0.5cm}
		\label{fig:icm}
	\end{center}
\end{figure}

Introducing variations to an empirical distribution can be as na\"ive as drawing random subsets.
While this is not guaranteed to introduce non-negligible variations, 
\cref{fig:icm} shows a crafted toy setup that illustrates the effect of these variations on the effect generation mechanism, and the asymmetry revealed as a result.

While several previous works relied on this principle for causal discovery in bivariate systems, they either impose strict constraints on the data types (e.g. continuous data in regression-based approaches or identical data spaces for cause and effect), tend to show high sensitivity to the capacity of the chosen model class, or suffer from prohibitive computational complexities that renders them practically applicable only to certain (e.g., binary) data types.
In this current work, we address these limitation, propose a new cause-effect identification framework based on artificially generated variations.
The choice of the discrepancy measure along with the kernel embedding of the marginal distributions renders our framework applicable a variety of data types\footnote{That is, within the identifiability limitations of the ICM postulate as shall be discussed in \cref{sub:vcei-identifiability}.} (e.g., timeseries data) and offers a practical realization leveraging convex optimization tools.

The problem we address in this paper is identifying the causal structure of a bivariate system from a single observational setting.
To that end, we proposed a two-step variation-based causal discovery approach relying on convex optimization to introduce non-negligible variations and the
kernel-based MMD metric to quantify the impact of these variations. Our contribution can thus be summarized as:
\begin{enumerate}
	\item We introduce a new frame work bivariate causal discovery from observational settings.
	\item We propose a kernel-based method that is independent of the data types used.
	\item The framework entails an optimization problem that has been cast as a convex optimization problem.
\end{enumerate}


\newpage

\section{Preliminaries}
\label{sec:preliminaries}

\paragraph{Assumptions:}
\label{pg:assumptions}
We will consider a bivariate system $(x,y)$ for cause-effect inference from an observational setting.
In such a system, we assume acyclicity and the existence of a causal link (i.e. either $x\to y$ or $y\to x$).
We additionally assume \emph{causal sufficiency} in the sense that all relevant covariates are observed.

\paragraph{Independence of Causal Mechanisms (ICM):}
\label{pg:icm}
Our identification framework relies principally on the ICM concept \cite{sgouritsa2016inference, peters2017elements} which postulates that the genuine data generation process decomposes into \emph{independent} modules that neither inform nor influence each other.
Such independence will not necessarily (and is in practice less likely to) hold in acausal decompositions.
In a bivariate causal graph $x\to y$ with a joint distribution $p_{xy}$, ICM implies 
\emph{independence} between the marginal $p_x$ and the conditional $p_{y|x}$, and shall be henceforth denoted by $p_{y|x} \perp p_x$.
ICM induces an asymmetry in bivariate systems that has been leveraged in several causal inference approaches \cite{mooij2009regression,janzing2010causal,stegle2010probabilistic,janzing2012information,daniusis2012inferring,scholkopf2012causal,sgouritsa2016inference,kocaoglu2017entropic,marx2017telling,tagasovska2018distinguishing,blobaum2018cause,budhathoki2018origo,marx2021formally}.
\citet{janzing2010causal} formulated this notion of \emph{independence} in terms of Kolmogorov complexities \cite{kolmogorov1968three} of the constituent distributions.
Many works thereafter relied on the \underline{m}inimum \underline{d}escription \underline{l}ength (MDL) \cite{rissanen1978modeling} as a proxy for the intractable Kolmogorov complexity \cite{budhathoki2017mdl,budhathoki2018origo,marx2018causal,mitrovic2018causal,tagasovska2018distinguishing,kalainathan2019generative,marx2019identifiability}.

\paragraph{Maximum Mean Discrepancy (MMD):}
\label{pg:mmd}
For analytical tractability, we will mainly consider kernel-based MMD as a metric of disparity between distributions \cite{gretton2008kernel,gretton2012kernel}.
Given a kernel $k$, the MMD can be expressed as norm in a \underline{r}eproducing \underline{k}ernel \underline{H}ilbert \underline{s}pace (RKHS) $\mathcal{H}$ between the kernel embeddings of the distributions $p$ and $q$:
\begin{align}
\text{MMD}^2_k(p, q) &= \left \| \mu_p  - \mu_q \right \|^2_{\mathcal{H}} 
\label{eq:mmd:norm}  
\end{align}
where $\mu_p$ and $\mu_q$ are the mean embeddings of $p$ and $q$, respectively, in the Hilbert space $\mathcal{H}$ through the feature mapping $k(x, \cdot)$.
From a practical perspective, squared MMD has an analytically tractable empirical estimator of a quadratic form given by:
\begin{equation}
\text{MMD}_k^2(p, q) \simeq   
                       \frac{1}{N^2} \sum_{i,j=1}^{N}   k(x_i, x_j)
					 - \frac{2}{NM}  \sum_{i,j=1}^{N,M} k(x_i, y_j) 
					 + \frac{1}{M^2} \sum_{i,j=1}^{M}   k(y_i, y_j)
\label{eq:mmd:empirical:biased}
\end{equation}
with $\{x_i\}_{i=1}^N$ and $\{y_i\}_{i=1}^M$ being finite sample sets drawn from $p$ and $q$, respectively \cite{sriperumbudur2009integral,gretton2012kernel}.
This efficient estimator renders MMD practically appealing for various applications amongst which is causal discovery \cite{goudet2017learning,baumann2020identifying,gao2021dag}.

\section{Variation-based Cause Effect Identification}
\label{sec:method}

In this section, we introduce our \underline{v}ariation-based \underline{c}ause \underline{e}ffect \underline{i}dentification (VCEI) framework, a two-step procedure performed at least once in each direction of a bivariate system to infer the genuine causal structure from a single observational setting.
Hypothesizing that the underlying causal structure is $x\to y$, the first step of VCEI is to introduce artificial variations to the marginal distribution $p_x$ (see \cref{sub:artificial-setups}).
In the second step, we quantify the impact of these variations on the conditional $p_{y|x}$ (see \cref{sub:quantifying-impact}).
According to the ICM postulate, variations on $p_x$ are expected to have minimal impact on the conditional $p_{y|x}$ in the genuine causal direction.

\paragraph{Notation:} let $\mathcal{D}=\{({x}_n, {y}_n)\}_{n=1}^N$ denote a set of $N$ i.i.d samples passively obtained, i.e. in an observational setting $p_{xy}$, from a bivariate system, where $x\in\mathbb{X}$ and $y\in\mathbb{Y}$ are two random variables following the marginals $p_x$ and $p_y$, respectively.
Let further $\mathcal{D}_{{x}}=\{{x}_n\,|\,({x}_n, {y}_n) \in \mathcal{D}\}$ denote the $x$-covariate view of the dataset, and likewise for $\mathcal{D}_y$.

\subsection{Artificially Generated Experimental Setups}
\label{sub:artificial-setups}

In this step, we propose an approach to introduce variations to the marginal distributions.
For simplicity though, we will describe our approach for the first random variable $x$, but it should be clear that this step takes place once for each covariate.
It should also be noted that such variations are intended to reveal potential dependencies between the marginal and the corresponding conditional, and do not necessarily retain similar dynamics to an \emph{intervention}.

Given $\mathcal{D}_x$ with their unknown marginal $p_x$, we define the \emph{empirical distribution} on these samples to be the uniform mixture of the Dirac delta distributions $\delta_{x_n}$ defined on each sample individually:
\begin{equation}
p_{x,N}({x}) = \frac{1}{N} \sum_{n=1}^N \delta({x}-{x}_n) = \frac{1}{N}\sum_{n=1}^N \delta_{{x}_n}(x)
\end{equation}
which is a probability density function with the corresponding empirical cumulative distribution function $F_{x,N}(x)$ (eCDF) defined on the sample set as
$
F_{x,N}({x}) = \frac{1}{N} \sum_{n=1}^{N} \mathds{1}_{{x}_n \,\leq\, {x}} 
$
where $\mathds{1}_{(\cdot)}$ is the indicator function and the inequality is to be understood entry-wise \citep{scott1992multivariate}.
A generalization of the empirical distribution is a weighed mixture of the constituent Dirac distributions $\delta_{x_n}$ which we will denote by $p_{x,N}^{\bm{\alpha}}$ and define as (see \cref{appendix:empirical-distribution} for a brief discussion on this modelling choice):
\begin{equation}
p_{x,N}^{\bm{\alpha}}({x}) = \sum_{n=1}^N \alpha_n\delta_{{x}_n}
\end{equation}
where $\bm{\alpha} = [\alpha_n]_{n=1}^N \in [0,1]^{N\times1}$ is a non-negative weight vector satisfying
$\bm{1}^\top \bm{\alpha} = 1$ where $\bm{1}$ is the all-ones vector.
From \cref{eq:mmd:empirical:biased}, the MMD between the empirical distribution $p_{x,N}$ and the weighted version thereof $p_{x,N}^{\bm{\alpha}}$ becomes:
\begin{equation}
\text{MMD}_k^2(p_{x,N}^{\bm{\alpha}},p_{x,N}) \simeq 
  \bm{\alpha}^\top \mathbf{K}_{xx} \bm{\alpha} 
- \frac{2}{N} \bm{\alpha}^\top \mathbf{K}_{xx} \bm{1}
+ \frac{1}{N^2} \bm{1}^\top \mathbf{K}_{xx} \bm{1}
\label{eq:mmd:weighted:estimate}
\end{equation}
where $\mathbf{K}_{xx}=[k(x_i,x_j)]_{i,j=1}^N$ is the Gram matrix of the kernel $k$ on the sample set $\mathcal{D}_x$.

With this defined, and with the objective of introducing a non-negligible variation to the marginal of ${x}$, we are interested in solving the following problem:

\paragraph{Problem 1}\label{prg:problem_1}\emph{Given a set of samples $\{{x}_n\}_{n=1}^N$, find the weight vector $\bm{\alpha}$ that renders the mixture distribution $p_{x,N}^{\bm{\alpha}}$ maximally distinct from $p_{x,N}$ in some discrepancy measure $D(\cdot, \cdot)$.}

For analytical tractability, we will mainly consider the (MMD) metric\footnote{While we introduce our framework based on the MMD metric, similar relaxations or heuristics \cite{park2017general} can be applied to render Problem 1 a convex optimization problem for other discrepancy measures. This is, however, outside the scope of this contribution.} w.r.t a positive definite kernel function $k_{\mathbb{X}}:\mathbb{X}^2\rightarrow\mathbb{R}$.
By adopting a kernel-based approach, we mask the data space (in the sense that data space, along with its type and dimensionality, is subsumed in the kernel design/function) with an appropriately chosen kernel $k_\mathbb{X}$ function rendering our VCEI framework widely applicable to various data types\footnote{For instance, in inferring summary graphs of temporal data using a timeseries kernel, or an embedding+kernel design for e.g. natural languages.} as opposed to e.g., regression-based identification frameworks.

Based on the squared MMD as a discrepancy measure, Problem 1 can be formally stated as:
\begin{align}
~~\underset{\bm{\alpha}}{\text{maximize}} ~~~~~ & \text{MMD}^2_{k_\mathbb{X}}(p_{x,N}^{\bm{\alpha}},\,p_{x,N}) 
												  \label{eq:original-formulation:objective}\\
                        \text{subject to} ~~~~~ & \bm{1}^\top \bm{\alpha} = 1 
                        						  \label{eq:original-formulation:equality}\\
                           					    & \bm{\alpha} \geqslant 0 \;\; \text{(entry-wise)
                           					      \label{eq:original-formulation:inequality}}
\end{align}

In spite of convexity of the objective (since MMD is jointly convex in both arguments as can be deduced from \cref{eq:mmd:norm}) and linearity of both constraints, the optimization problem remains non-convex.
This is due the fact that the convex objective is being maximized rather than minimized which renders the objective a concave function in the standard form of a convex optimization problem.

Noting that the closed-form estimator of the squared MMD is also quadratic in the optimization variable $\bm{\alpha}$ (see \cref{eq:mmd:weighted:estimate}), \citet{park2017general} address this problem in a two-step procedure referred to as \underline{s}emi\underline{d}efinite \underline{r}elaxation (SDR). 
They first \emph{lift} the problem to a higher dimensional space by defining $\mathbf{A}=\bm{\alpha}\bm{\alpha}^\top$ in which the objective function becomes linear, then apply a convex \emph{relaxation} to the intractable constraints.
As a result, the following formulation is a relaxation of \ref{eq:original-formulation:equality}--\ref{eq:original-formulation:inequality} (see \cref{appendix:derivation}for a derivation) which is a \underline{q}uadratically \underline{c}onstraint \underline{q}uadratic \underline{p}rogram (QCQP) that can make use of off-the-shelf convex optimization tools\footnote{For instance, we used the open-source library \texttt{cvxpy} \cite{diamond2016cvxpy} for all experiments.}:
\begin{align}
~~\underset{\mathbf{A}}{\text{maximize}} ~~~~~ &  \mathbf{A} \bullet \left(\mathbf{K}_{xx}-\frac{2}{N}\mathbf{K}_{xx}\bm{1}\bm{1}^\top \right) + \frac{1}{N^2} \bm{1}^\top\mathbf{K}_{xx}\bm{1} 
		\label{eq:sdr-formulatoin:objective}   \\
                       \text{subject to} ~~~~~ &  \begin{bmatrix}
													\mathbf{A} 					& \mathbf{A}\bm{1} \\
													\bm{1}^\top\mathbf{A}		& 1 \\
                       							  \end{bmatrix} \;\succeq \;0   \quad \text{(positive semidefiniteness)}
        \label{eq:sdr-formulation:inequality:psd}  \\
        									   & \mathbf{A} \geqslant 0 ~\qquad\qquad\qquad \text{(entry-wise)}
        \label{eq:sdr-formulation:inequality:entry} \\
        									   & \bm{1}^\top\mathbf{A}\bm{1} = 1 
        \label{eq:sdr-formulation:equality:normalization} \\
        									   & \mathbf{A} = \mathbf{A}^\top 
        \label{eq:sdr-formulation:equality:symmetry}      									   
\end{align}
where $\mathbf{K}_{xx}=\left[k_{\mathbb{X}}({x}, \tilde{{x}})\right]_{x,\tilde{x}\in\mathcal{D}_x}$ is the Gram matrix, and $\bullet$ denotes the dot-product in matrix space defined as $\mathbf{A}\bullet\mathbf{K}_{xx} = \text{\textbf{trace}}(\mathbf{A}\mathbf{K}_{xx})$.

The solution $\mathbf{A}^{\text{SDR}}$ to \ref{eq:sdr-formulatoin:objective}--\ref{eq:sdr-formulation:equality:symmetry} is an optimal solution to the original formulation $\mathbf{A}^\star$ \ref{eq:original-formulation:objective}-\ref{eq:original-formulation:inequality} (i.e. $\mathbf{A}^{\text{SDR}} \equiv \mathbf{A}^\star$) if the condition $\mathbf{A}^\star=\bm{\alpha}^\star\bm{\alpha}^{\star\top}$ is satisfied (i.e. if $\mathbf{A}^\text{SDR}$ is rank one which will be the case if $\mathbf{A}^{\text{SDR}}$ is a feasible solution to  \ref{eq:original-formulation:objective}-\ref{eq:original-formulation:inequality} \cite{park2017general})\footnote{In \cref{sub:practical-considerations}, we discuss situations in which $\mathbf{A}^{\text{SDR}}$ is not a rank one matrix.}.
In this case, the distribution weights can be recovered as $\bm{\alpha}^\star=\mathbf{A}^\star\bm{1}$.

With the solution to Problem 1, we would have obtained a new marginal $p_{x,N}^{\bm{\alpha}^\star}$ that is constructed from the passively obtained observational data $\mathcal{D}_x$ and is maximally distinct from the original marginal $p_x$.
Finally, this optimization is performed on the second covariate $y$ to obtain a weighted marginal $p_{y,N}^{\bm{\beta}}$ with weight vector $\bm{\beta}\in[0,1]^{N\times1}$ that is maximally distinct from $p_{y,N}$.

\subsection{Quantifying the Impact of Distributional Variations}
\label{sub:quantifying-impact}

In the second step, we quantify the impact of the artificially generated variations (i.e. within the marginals $p_{x,N}$ and $p_{x, N}^{\bm{\alpha}}$ and similarly from $p_{y,N}$ to $p_{y, N}^{\bm{\beta}}$) on the conditionals $p_{x|y}$ and $p_{y|x}$, respectively.
This can be achieved by fitting predictive models to each of these settings leading to the two models $\hat{f}_{y|x}$ and $\hat{f}_{y|x}^{\bm{\alpha}}$ in the $x\to y$ direction, and $\hat{g}_{x|y}$ and $\hat{g}_{x|y}^{\bm{\beta}}$ in the opposite direction.
Each model is attainable from a model class $\mathcal{M}_{x\to y}$ or $\mathcal{M}_{y\to y}$ with their corresponding training paradigms $\text{Train}_{\mathcal{M}_{x\to y}}[\cdot]$ and $\text{Train}_{\mathcal{M}_{y\to x}}[\cdot]$.

In order to fit a predictive model on a weighted empirical distribution e.g. $p_{x,N}^{\bm{\alpha}}$, the corresponding weights can be considered sample weights and the training paradigms $\text{Train}_{\mathcal{M}_\cdot}[\cdot]$ supports sample importance\footnote{Alternatively, model fitting can be preceded by a re-sampling step.} (see, for example, \cite{wen2018weighted} for a weighted Gaussian Process (GP) model or \cite{steininger2021density} for neural networks).

ICM postulates that, if $x\to y$ is the true causal direction of the data generation process, then the impact of the introduced variations on the $\hat{g}$ models are likely to be more apparent.
We quantify this impact via model disagreement on a (potentially unlabeld) set \cite{nakkiran2020distributional}, which is in turn quantified as the MMD discrepancy between each model's prediction on a common set:
\begin{equation}
S_{x\to y} = \text{MMD}_{k_\mathbb{Y}}^2
				\left(
					\hat{f}_{y|x}(x),
					\hat{f}_{y|x}^{\bm{\alpha}}(x)
				\right)
\end{equation}
where $x\sim p_x(x)$ (which empirically could simply be all samples in $\mathcal{D}_x$ or a random subset thereof)
and similarly for $S_{y\to x}$.
Finally, the lower of either scores\footnote{Similarly, see \cref{sub:practical-considerations} for a discussion the implicit assumptions this decision criterion entails.} $S_{x\to y}$ and $S_{y\to x}$ is an indicator of a lesser impact on the conditionals, and in turn the genuine causal direction.
An overview of the VCEI framework for identical data spaces is presented in \cref{alg:VCEI}.

\begin{algorithm}[!t]
	\caption{Variation-based cause-effect identification (VCEI) on identical data spaces $\mathbb{X}\equiv\mathbb{Y}$}
	\label{alg:VCEI}
	\begin{algorithmic}
		\REQUIRE $\mathcal{D}=\{(x_n, y_n)\}_{n=1}^N$, a kernel function $k$, model classes $\mathcal{M}_{x \to y}$, $\mathcal{M}_{y \to x}$, corresponding training paradigms $\text{Train}_{\mathcal{M}_{x\to y}}[\cdot]$ and $\text{Train}_{\mathcal{M}_{y\to x}}[\cdot]$, and a regularization parameter $b_{\alpha}$. 
		\ENSURE $\mathbb{X} \equiv \mathbb{Y}$ (where $x\in\mathbb{X}$ and $y\in\mathbb{Y}$)
		\STATE \textbf{Estimate }$S_{x\to y}$\textbf{:} Solve SDR of Problem 1 (\Cref{eq:sdr-formulatoin:objective}--\ref{eq:sdr-formulation:equality:symmetry} and \ref{eq:augment:alpha-max}) in $\mathcal{D}_x$ to estimate $\bm{\alpha}$
		\STATE $\hat{f}_{y|x} \leftarrow \text{Train}_{\mathcal{M}_{x \to y}}\left[p_{xy,N}\right]$
		\STATE $\hat{f}_{y|x}^{\bm{\alpha}} \leftarrow \text{Train}_{\mathcal{M}_{x \to y}}\left[p_{xy,N}^{\bm{\alpha}}\right]$
		\STATE $S_{x\to y} \leftarrow \text{MMD}_k^2\left(\hat{f}_{y|x}(p_{x,N}), \hat{f}_{y|x}^{\bm{\alpha}}(p_{x,N})\right)$
		\STATE \textbf{Estimate }$S_{y\to x}$\textbf{:} Solve SDR of Problem 1 (\Cref{eq:sdr-formulatoin:objective}--\ref{eq:sdr-formulation:equality:symmetry} and \ref{eq:augment:alpha-max}) in $\mathcal{D}_y$ to estimate $\bm{\beta}$
		\STATE $\hat{g}_{x|y} 			    \leftarrow \text{Train}_{\mathcal{M}_{y\to x}}\left[p_{xy,N}\right]$
		\STATE $\hat{g}_{x|y}^{\bm{\beta}}  \leftarrow \text{Train}_{\mathcal{M}_{y\to x}}\left[p_{xy,N}^{\bm{\beta}}\right]$
		\STATE $S_{y\to x} 					\leftarrow \text{MMD}_k^2\left(\hat{g}_{x|y}(p_{y,N}), \hat{g}_{x|y}^{\bm{\beta}}(p_{y,N})\right)$		
		\STATE $~~~$\textbf{Return:} $``x\to y``$ \textbf{if} $S_{x\to y} < S_{y \to y}$ \textbf{otherwise} $``y\to x``$
		
	\end{algorithmic}
\end{algorithm}

\subsection{Practical Considerations}
\label{sub:practical-considerations}

While Problem 1 tends to construct setups with maximal disparity from the given empirical distribution $p_{x,N}$, we are not necessarily interested in such extreme scenarios as long as these variations are non-negligible so that they reveal dependencies between the marginal and the conditional distributions in the acausal direction.
Therefore, we would oftentimes prefer a sub-optimal, yet more appealing, solution to the optimal solution of Problem 1 for practical considerations.
Such practical aspects are discussed in the sequel.

\textbf{Scalability:} one directly notes that the SDR formulation \ref{eq:sdr-formulatoin:objective}--\ref{eq:sdr-formulation:equality:symmetry} hardly scales to larger datasets since the dimensionality of the optimization space is quadratic in the number of data points $N$ (as a result of the lifting step).
Therefore, we rather restrict the weighted distribution $p_{\cdot,N}^{\bm{\alpha}}$ to a reasonable number of samples $M<N$ drawn randomly from the original dataset.
This is denoted henceforth by $p_{\cdot,M}$ for the $M$-sample subset and $p_{\cdot,M}^{\tilde{\bm{\alpha}}}$ for the weighted version thereof.
The size of the reference empirical distribution $p_{\cdot,N}$ (2\textsuperscript{nd} argument of \cref{eq:mmd:weighted:estimate}) does not affect the dimensionality of the optimization problem and, thus, can grow as needed within the Gram matrix computational limits.

\begin{wrapfigure}[21]{r}{0.45\textwidth}
	\centering
	\vspace{-0.55cm}
	\includegraphics[width=6.2cm]{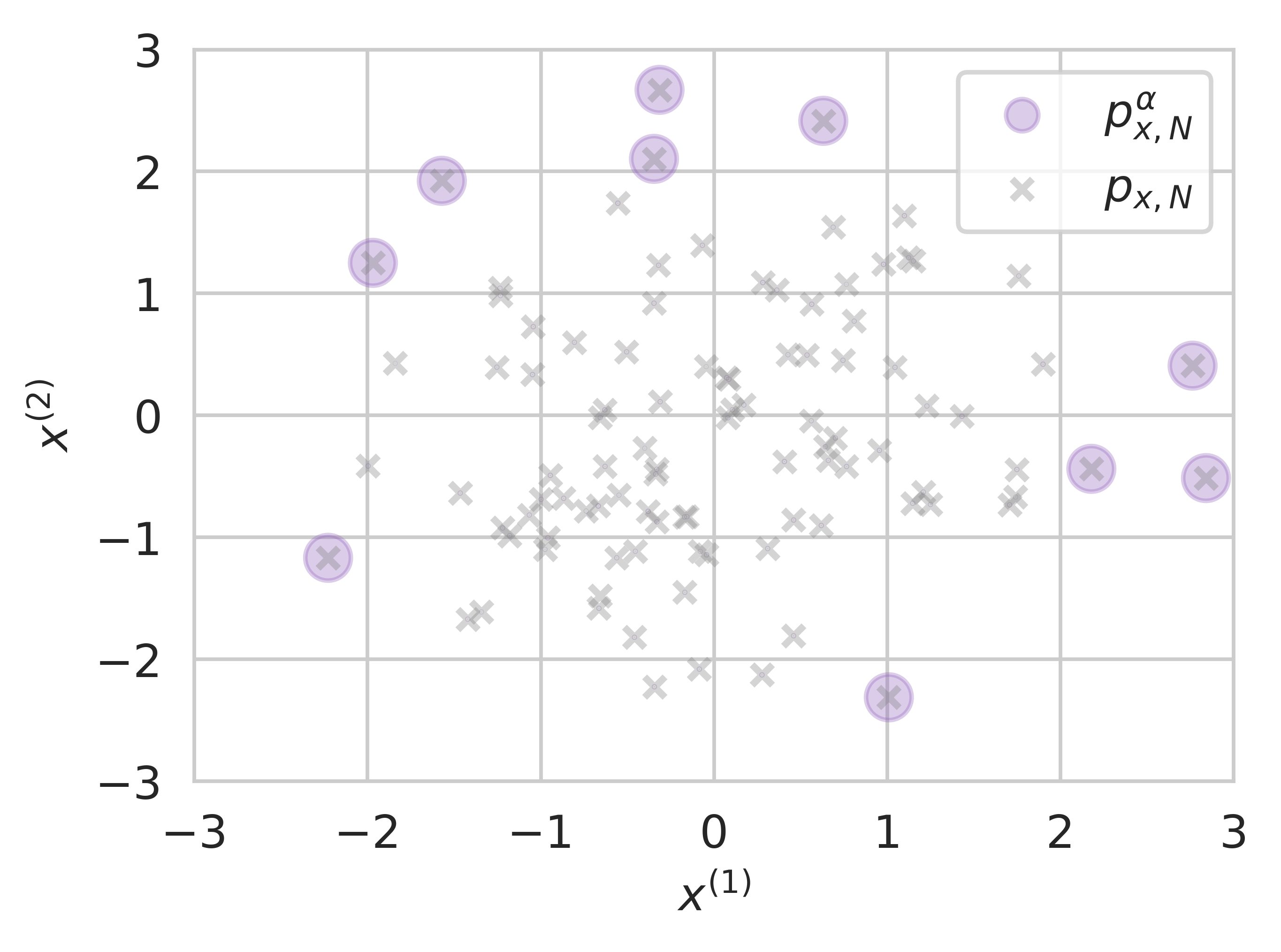}
	\vspace{-0.1cm}
	\caption{An illustrative example of solving problem 1 on a 2D Gaussian dataset.
		The true distribution is $p_x=\mathcal{N}(\bm{0}, \bm{1})$ from which $N=100$ samples are depicted in grey.
		Purple markers represent the weights $\bm{\alpha}$ of the weighted distribution $p_{x,100}^{\bm{\alpha}}$.
		}
	\label{fig:outliers}
\end{wrapfigure}

\textbf{Dirac Distributions:} an artifact of the choice of the discrepancy measure (and the formulation of problem 1) is that attainable solutions to \ref{eq:sdr-formulatoin:objective}--\ref{eq:sdr-formulation:equality:symmetry} are in practice Dirac-like probability measures in the sense that $\left\|\bm{\alpha}\right\|_\infty\sim1$ where $\left\|\cdot\right\|_\infty$ is the supremum norm.
One can avoid such extreme scenarios by augmenting the optimization problem with regularizing constraints such as
\begin{align}
\left\| \mathbf{A} \right\|_\infty \leqslant b_{\alpha}
\label{eq:augment:alpha-max}
\end{align}
with the supremum norm of a matrix given by $\left\| \mathbf{A} \right\|_\infty \coloneqq \max_i \left\| \mathbf{a}_{i\cdot}\right\|_1$
which directly constraints the maximum probability mass that is allowed on a single data point
and $b_\alpha\in[1/M,\,1.0]$ becomes a hyper-parameter in our framework. \Cref{fig:outliers} illustrates the effect of this regularization constraint on a 2D sample set drawn from a standard Normal distribution. 
Likewise, one can constrain maximum deviation from the uniform mixture as in
\begin{equation}
\text{MMD}_k^2\left(p_{\cdot,M}^{\tilde{\bm{\alpha}}}, p_{\cdot, M}\right)
\;\leqslant\;
\text{MMD}^2\left(p_{\cdot,M}, p_{\cdot,N}\right) + b_D
\label{eq:augment:mmd-uniform-max}
\end{equation}
where $b_D$ is a slack variable, and the l.h.s is a linear function of the optimization variable $\mathbf{A}$ similar to Eq. \ref{eq:sdr-formulatoin:objective} with a different Gram matrix.
Given the convexity of both regularization constraints above, \cref{eq:augment:alpha-max,eq:augment:mmd-uniform-max}, the SDR formulation \ref{eq:sdr-formulatoin:objective}--\ref{eq:sdr-formulation:equality:symmetry} remains a convex optimization problem if augmented wither either of these constraints.

\textbf{SDR Relaxation:} a solution $d_\mathbb{X}^\text{sdr}$ obtained from the SDR formulation is a lower bound on the optimal value of the original formulation \ref{eq:original-formulation:objective}--\ref{eq:original-formulation:inequality} that is tight only if the rank one condition $\mathbf{A}=\bm{\alpha}\bm{\alpha}^\top$ is satisfied \cite{park2017general}. 
Yet, the rank-one condition is not guaranteed, and is even unlikely to be satisfied as additional constraints (e.g., \cref{eq:augment:alpha-max,eq:augment:mmd-uniform-max}) are included in the optimization problem.
Practically, however, estimating the weight vector as $\bm{\alpha} \simeq \mathbf{A}^{\text{SDR}}\bm{1}$ remains a reasonable estimate for the weighted empirical that notably outperforms naive baselines (e.g. drawing random subsets).

\textbf{Disagreement Bias:} 
in the second step of our identification framework, we quantify disparity between two models (e.g., $\hat{f}_{y|x}$ and $\hat{f}_{y|x}^{\bm{\alpha}}$) via their MMD-based disagreement on a common input distribution. 
However, for some model classes (e.g., neural networks) such an approach is likely to be biased.
In fact, it was observed recently that two identical neural network classifiers would disagree even when trained on identical data as long as a randomization factor plays a roll (i.e. different initial weights, batching, data shuffling, or different random seeds in general) \cite{nakkiran2020distributional,jiang2021assessing}.
In fact, it was conjectured  that this sort of disagreement correlates with the generalization performance of the classifier.

\begin{wrapfigure}[21]{r}{0.45\textwidth}
	\centering
	\vspace{-0.55cm}
	\includegraphics[width=6.2cm]{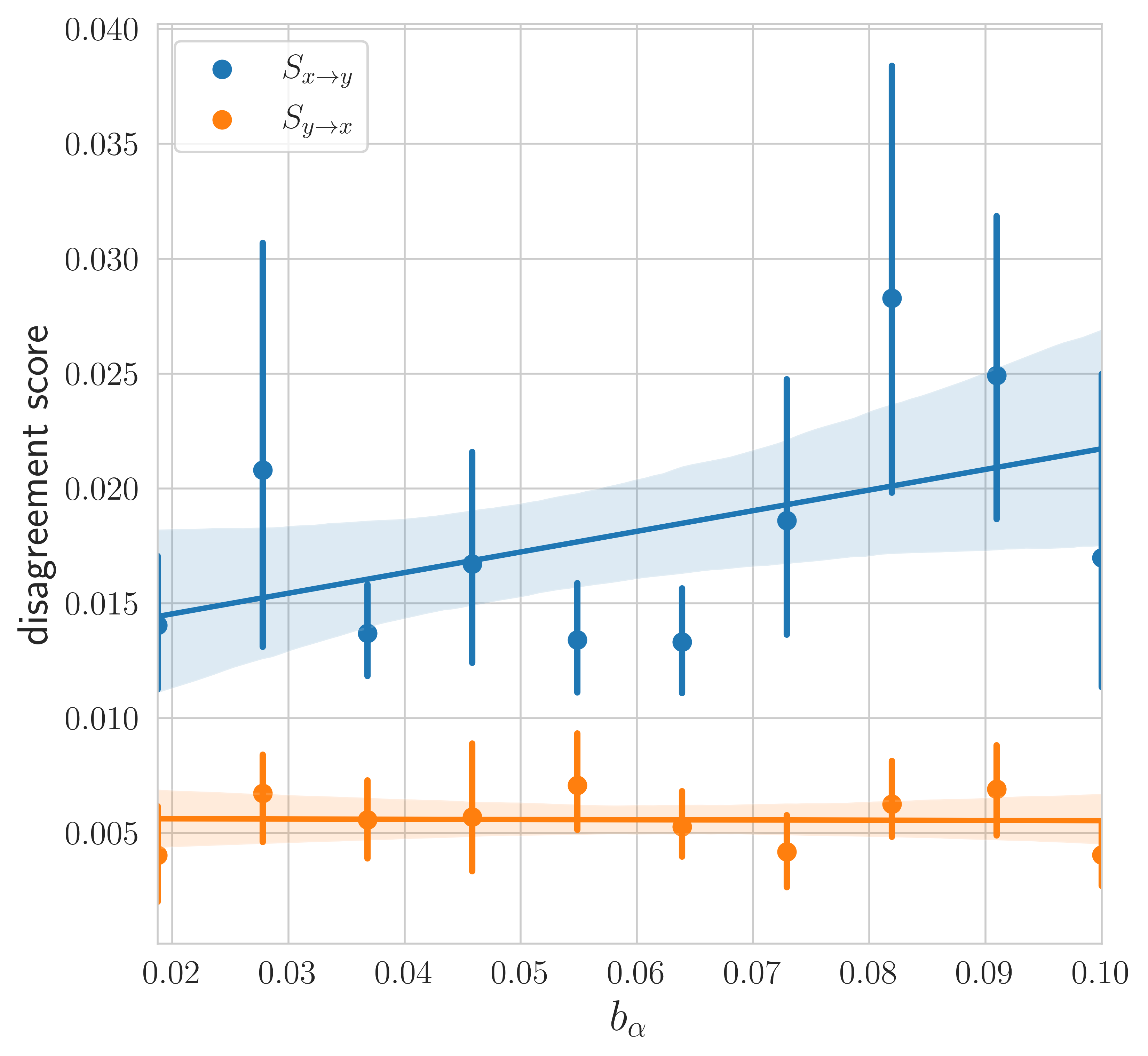}
	\vspace{-0.2cm}
	\caption{An illustration of the behaviour of the disagreement scores $S_{x\to y}$ (upper) and $S_{y\to x}$ (lower) for different values of the hyper-parameter $b_\alpha$ where the true causal structure is $y\to x$.
	Example from the 1\textsuperscript{st} pair of the \texttt{SIM} dataset \cite{cause_effect_moiij}.}
	\label{fig:trend}
\end{wrapfigure}

Our empirical observations extend the claims of \cite{nakkiran2020distributional,jiang2021assessing} to regression problems with MMD as a disagreement metric.
Since all our models are trained on limited data, they are likely to disagree (i.e. generalize poorly) even if the training distributions were identical.
This \emph{disagreement bias} is not accounted for in our work, and is left as an open question for future contribution.
\Cref{fig:trend} depicts an example of such a bias in the non-zero disagreement score $S_{y\to x}$ even though the genuine causal direction is indeed $y\to x$.

\textbf{Trend as a Score:}
the final decision criterion, that is comparing the MMD-based disagreements scores, implicitly imposes a strong assumption of the data spaces $\mathbb{X}\equiv\mathbb{Y}$ and similarly on the kernels $k_{\mathbb{X}}\equiv k_{\mathbb{Y}}$ (admittedly, this has been an implicit assumption in numerous previous works e.g., roughly all approaches relying on regression performance).
At the expense of additional computational demands, one can circumvent this limitation with the following observation.
It is observed, and also intuitive, that the attainable solution to \ref{eq:sdr-formulatoin:objective}--\ref{eq:sdr-formulation:equality:symmetry} augmented with \ref{eq:augment:alpha-max} is monotonic in the hyper-parameter $b_\alpha$ (refer to \cref{appendix:experiment-setup} for an illustrative example).
According to ICM, repeating the optimization problem with increasing values for $b_\alpha$ is likely to be reflected in an increasing trend of the disagreement score of the acausal direction.
In the causal direction, however, the disagreement score is expected to remain roughly constant.

Treating these disagreement score as functions of the regularization hyper-parameter (e.g., linearly regressing $S_{.}$ on $b_\alpha$ for different solutions of the optimization problem) gives an alternative decision criterion (e.g., trend of these regression lines) that is independent of the data spaces, kernels, and kernel hyper-parameters.
This is briefly illustrated in \cref{fig:trend} (and a similar effect can be observed w.r.t the number of samples $M$), but is not thoroughly investigated in this work, and is rather left as another open point for future contribution.
Interestingly, and also left open for future work, using this decision mechanism may also mitigate the causal sufficiency assumption leading to broader identifiability.

\subsection{Identifiability}
\label{sub:vcei-identifiability}

The proposed VCEI framework is viewed as a practical realization of the ICM principle, and thus, inherits all identifiability limitations of that postulate. 
When viewed from e.g., Kolmogorov complexities $K(p_x) + K(p_{y|x}) \leq K(p_y) + K(p_{x|y})$ if $x\to y$ as formulated by \citet{janzing2010causal}, one directly notes a limitation of ICM-based frameworks, that is when equality occurs and thus the ICM-based asymmetry vanishes.
Asymmetry vanishes if the underlying system can be described with the same functional form and distributional families in either direction \citep{mitrovic2018causal}. 
A very common example thereof are linear models with additive Gaussian noise\citep{hoyer2008nonlinear}.
Loosely speaking, identifiability of ICM-based frameworks increase with increasing non-linearity of the functional form, smaller noise effects \citep{mooij2016distinguishing}, and less (or no) confounding bias .

In addition, and as stated earlier, we assume existence of a causal link and causal sufficiency. 
The former, however, can be mitigated with either an independence test. 
The latter can also be mitigated with the use of disagreement trends rather than single scores (as discussed in the preceding subsection) where confounding may lead to a positive trend in either direction, but it is expected be more observable (i.e. steeper) in the acausal direction.

\section{Experimental Validation}
\label{sec:experiments}


In the sequel, we report empirical validation of our proposed method.
For a benchmark, we tested VCEI on the same use-cases presented in the work of \citet{tagasovska2018distinguishing}.

\textbf{Simulated data:} simulation data\footnote{All synthetic dataset have been obtained from: https://github.com/tagas/bQCD} were originally generated in the work of \citet{mooij2016distinguishing}.
Four different scenarios were considered: 
\texttt{SIM} which is the default use-case without confounder-bias, 
\texttt{SIM-c} which includes a single latent confounder, 
\texttt{SIM-ln} a use-case with low noise levels, and finally
\texttt{SIM-G} which has a Gaussian-like distribution for both the cause $X$ and the additive noise.
We additionally, included the 5 additional synthetic datasets published by \citet{tagasovska2018distinguishing} (namely \texttt{AN(-s)}, \texttt{LS(-s)}, and \texttt{MN-U}) (see \cref{appendix:sub:benchmark} for a more detailed description).

\textbf{Real-world data:} the Tübingen Cause-Effect (CE) benchmark was considered for real-data validation, which consists of 108 pairs from 37 different domains. We only used 103 pairs, which have univariate (continuous or discrete) cause and effect variables.

\textbf{Baselines:} we included a selected set of the methods from the baselines reported in the work of \citet{tagasovska2018distinguishing}.
We namely compare our VCEI framework to: 
 biCAM \cite{buhlmann2014cam}, which are \underline{a}dditive \underline{n}oise \underline{m}odel (ANM)-based,
 IGCI \cite{janzing2010causal}, 
 bQCD \cite{tagasovska2018distinguishing}),
 Sloppy \cite{marx2019identifiability}, and finally
 GPI \cite{stegle2010probabilistic}.


\textbf{Sample Size:} due to the limited scalability of the proposed framework (and the limited computational budget), the number of samples used in the optimization step to construct the different setting \ref{sub:artificial-setups}, and latter for training of the predictive models, was chosen to be relatively low.

\begin{figure}[t]
	\begin{center}
		\includegraphics[width=0.9\linewidth]{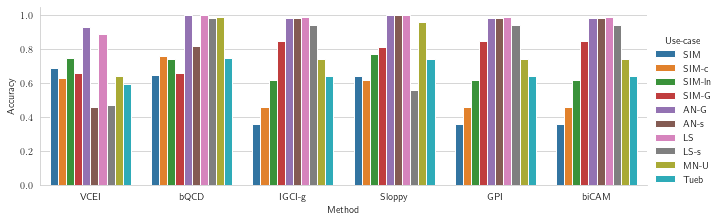}
		\caption{Accuracy of VCEI on benchmark datasets compared to baseline methods. Identification accuracies for baseline methods were taken from \cite{tagasovska2018distinguishing}. For \texttt{SIM-*} the sample size was $M=200$, while for the remaining datasets the sample was limited to $M=100$.}\label{fig:rel_acc}
		\vspace{-0.5cm}
	\end{center}
\end{figure}

\Cref{fig:rel_acc} depicts the identification accuracies of our method on the selected benchmark datasets, and compared to other causal discovery baseline algorithms. 
We use the same metric as  in \cite{mooij2016distinguishing} namely, \emph{accuracy for forced decisions}.
In principle, each algorithm is forced to take a decision about the causal direction from which the identification accuracy corresponds to the how frequent the algorithm was able to reach correct decisions over the number of dataset files.

While our framework does not show an unprecedented performance on the benchmark dataset, it is certainly competitive to many previous methods, in addition to being generic w.r.t data types, and robust w.r.t choice of model class and the learning capacity thereof.

\section{Conclusion}
\label{sec:conclusion}

We introduce Variation-based Cause Effect Identification (VCEI), a kernel-based framework for causal discovery in a bivariate systems.
Our method combines the principle of independent causal mechanism (ICM) with convex-optimization under semi-definite relaxation (SDR) and the learning power of data-driven models to identify the genuine causal structure of a bivariate system. 
With the kernel-based scores, we impose only mild assumptions on the the data types, thus giving the advantage of its implementation for a wide range of applications.
Additionally, our framework is robust to the model capacity as long as it is capacitive enough to learn variations of conditionals.

\section{Related work}
\label{sec:related-work}

In this section, we briefly review relevant work on causal discovery in bivariate systems.
The intent is not to provide an extensive review (for which the interested reader is referred to e.g. \cite{mooij2016distinguishing} specifically for cause-effect identification or \cite{vowels2021d} for a more recent review on causal discovery).
Rather, we review works that notably share similarities and analogies to our proposed framework in order to highlight and emphasize our contributions.

Works on causal discovery started with conditional independence tests \cite{spirtes2000causation,sun2007distinguishing,pearl2009causality} which fell short in bivariate cause-effect identification scenarios due to lack of conditioning covariates.
Lines of work that addressed this problem postulated a sort of an inherent asymmetry in the cause-effect relationship.
An example of such is the functional and distributional asymmetries proposed by the early works in this direction \cite{shimizu2006linear,hoyer2008nonlinear,mooij2009regression,zhang2012identifiability}.
Contrary to these frameworks, our proposed approach does not impose functional or distributional constraints on the causal relationship.

A different aspect of asymmetry is the ICM postulate on which numerous cause-effect identification frameworks have relied  \cite{sgouritsa2016inference,mooij2009regression,janzing2010causal,stegle2010probabilistic,janzing2012information,daniusis2012inferring,scholkopf2012causal,kocaoglu2017entropic,marx2017telling,tagasovska2018distinguishing,blobaum2018cause,budhathoki2018origo,marx2021formally,budhathoki2017mdl,kalainathan2019generative,marx2018causal,marx2019identifiability,mitrovic2018causal}, mainly utilizing the MDL as a proxy in place of the intractable Kolmogorov complexities.
Yet, most of the works are limited specific data spaces, e.g. numeric data for regression-based frameworks \cite{sgouritsa2016inference,mooij2009regression,tagasovska2018distinguishing,marx2019identifiability}.
Notable exceptions are works relying on kernel-embeddings \cite{mitrovic2018causal,lopez2015towards}
Likewise, our contribution lifts all constraints on the data spaces via the adopted kernel-based MMD metric (except for a mild assumption discussed in \cref{sub:practical-considerations}) to the choice of a characteristic kernel.

Kernel-based MMD was utilized a loss function in \cite{goudet2017learning} for learning bivariate causal structures. Their approach relies on the simplicity of the functional relationship in the causal direction, and thus can be identified with a model class of limited-capacity.
The higher the model capacity, the less identifiable a causal structure would be to their model.
In contrast, our framework is more robust to the model choice in the sense that it only requires a model class of a capacitive power to learn the functional relationship in either direction equally well.

Finally, and aside from bivariate systems, \citet{peters2016causal} proposed a causal discovery framework in scenarios of multiple experimental setups (including an observational one) with random, unknown interventions.
Yet, in the case of a single observational setup, they introduce conditional splitting of the dataset (under predefined conditions) to emulate an artificial scenario of multiple experimental setups.
In spite of the distinction, their contribution was an inspiration for our proposed framework.

\newpage
\bibliography{references}

\begin{thebibliography}{46}
\providecommand{\natexlab}[1]{#1}
\providecommand{\url}[1]{\texttt{#1}}
\expandafter\ifx\csname urlstyle\endcsname\relax
  \providecommand{\doi}[1]{doi: #1}\else
  \providecommand{\doi}{doi: \begingroup \urlstyle{rm}\Url}\fi

\bibitem[Baumann et~al.(2020)Baumann, Solowjow, Johansson, and
  Trimpe]{baumann2020identifying}
Dominik Baumann, Friedrich Solowjow, Karl~H Johansson, and Sebastian Trimpe.
\newblock Identifying causal structure in dynamical systems.
\newblock \emph{arXiv preprint arXiv:2006.03906}, 2020.

\bibitem[Bl{\"o}baum et~al.(2018)Bl{\"o}baum, Janzing, Washio, Shimizu, and
  Sch{\"o}lkopf]{blobaum2018cause}
Patrick Bl{\"o}baum, Dominik Janzing, Takashi Washio, Shohei Shimizu, and
  Bernhard Sch{\"o}lkopf.
\newblock Cause-effect inference by comparing regression errors.
\newblock In \emph{International Conference on Artificial Intelligence and
  Statistics}, pp.\  900--909. PMLR, 2018.

\bibitem[Budhathoki \& Vreeken(2017)Budhathoki and Vreeken]{budhathoki2017mdl}
Kailash Budhathoki and Jilles Vreeken.
\newblock Mdl for causal inference on discrete data.
\newblock In \emph{2017 IEEE International Conference on Data Mining (ICDM)},
  pp.\  751--756. IEEE, 2017.

\bibitem[Budhathoki \& Vreeken(2018)Budhathoki and
  Vreeken]{budhathoki2018origo}
Kailash Budhathoki and Jilles Vreeken.
\newblock Origo: causal inference by compression.
\newblock \emph{Knowledge and Information Systems}, 56\penalty0 (2):\penalty0
  285--307, 2018.

\bibitem[B{\"u}hlmann et~al.(2014)B{\"u}hlmann, Peters, and
  Ernest]{buhlmann2014cam}
Peter B{\"u}hlmann, Jonas Peters, and Jan Ernest.
\newblock {CAM: Causal additive models, high-dimensional order search and
  penalized regression}.
\newblock \emph{The Annals of Statistics}, 42\penalty0 (6):\penalty0
  2526--2556, 2014.

\bibitem[Daniusis et~al.(2012)Daniusis, Janzing, Mooij, Zscheischler, Steudel,
  Zhang, and Sch{\"o}lkopf]{daniusis2012inferring}
Povilas Daniusis, Dominik Janzing, Joris Mooij, Jakob Zscheischler, Bastian
  Steudel, Kun Zhang, and Bernhard Sch{\"o}lkopf.
\newblock Inferring deterministic causal relations.
\newblock \emph{arXiv preprint arXiv:1203.3475}, 2012.

\bibitem[Diamond \& Boyd(2016)Diamond and Boyd]{diamond2016cvxpy}
Steven Diamond and Stephen Boyd.
\newblock Cvxpy: A python-embedded modeling language for convex optimization.
\newblock \emph{The Journal of Machine Learning Research}, 17\penalty0
  (1):\penalty0 2909--2913, 2016.

\bibitem[Gao et~al.(2021)Gao, Shen, and Xia]{gao2021dag}
Yinghua Gao, Li~Shen, and Shu-Tao Xia.
\newblock {DAG-GAN: Causal Structure Learning with Generative Adversarial
  Nets}.
\newblock In \emph{ICASSP 2021-2021 IEEE International Conference on Acoustics,
  Speech and Signal Processing (ICASSP)}, pp.\  3320--3324. IEEE, 2021.

\bibitem[Goudet et~al.(2017)Goudet, Kalainathan, Caillou, Guyon, Lopez-Paz, and
  Sebag]{goudet2017learning}
Olivier Goudet, Diviyan Kalainathan, Philippe Caillou, Isabelle Guyon, David
  Lopez-Paz, and Michèle Sebag.
\newblock {Learning Functional Causal Models with Generative Neural Networks}.
\newblock \emph{arXiv.org}, September 2017.
\newblock \doi{10.1007/978-3-319-98131-4, 10.48550/arXiv.1709.05321}.
\newblock URL \url{https://arxiv.org/abs/1709.05321v3}.

\bibitem[Gretton et~al.(2008)Gretton, Borgwardt, Rasch, Scholkopf, and
  Smola]{gretton2008kernel}
Arthur Gretton, Karsten Borgwardt, Malte~J Rasch, Bernhard Scholkopf, and
  Alexander~J Smola.
\newblock A kernel method for the two-sample problem.
\newblock \emph{arXiv preprint arXiv:0805.2368}, 2008.

\bibitem[Gretton et~al.(2012)Gretton, Borgwardt, Rasch, Sch{\"o}lkopf, and
  Smola]{gretton2012kernel}
Arthur Gretton, Karsten~M Borgwardt, Malte~J Rasch, Bernhard Sch{\"o}lkopf, and
  Alexander Smola.
\newblock {A Kernel Two-Sample Test}.
\newblock \emph{The Journal of Machine Learning Research}, 13\penalty0
  (1):\penalty0 723--773, 2012.

\bibitem[Hoyer et~al.(2008)Hoyer, Janzing, Mooij, Peters, and
  Sch{\"o}lkopf]{hoyer2008nonlinear}
Patrik Hoyer, Dominik Janzing, Joris~M Mooij, Jonas Peters, and Bernhard
  Sch{\"o}lkopf.
\newblock Nonlinear causal discovery with additive noise models.
\newblock \emph{Advances in neural information processing systems}, 21, 2008.

\bibitem[Janzing \& Sch{\"o}lkopf(2010)Janzing and
  Sch{\"o}lkopf]{janzing2010causal}
Dominik Janzing and Bernhard Sch{\"o}lkopf.
\newblock {Causal inference using the algorithmic Markov condition}.
\newblock \emph{IEEE Transactions on Information Theory}, 56\penalty0
  (10):\penalty0 5168--5194, 2010.

\bibitem[Janzing et~al.(2012)Janzing, Mooij, Zhang, Lemeire, Zscheischler,
  Daniu{\v{s}}is, Steudel, and Sch{\"o}lkopf]{janzing2012information}
Dominik Janzing, Joris Mooij, Kun Zhang, Jan Lemeire, Jakob Zscheischler,
  Povilas Daniu{\v{s}}is, Bastian Steudel, and Bernhard Sch{\"o}lkopf.
\newblock Information-geometric approach to inferring causal directions.
\newblock \emph{Artificial Intelligence}, 182:\penalty0 1--31, 2012.

\bibitem[Jiang et~al.(2021)Jiang, Nagarajan, Baek, and
  Kolter]{jiang2021assessing}
Yiding Jiang, Vaishnavh Nagarajan, Christina Baek, and J~Zico Kolter.
\newblock Assessing generalization of sgd via disagreement.
\newblock \emph{arXiv preprint arXiv:2106.13799}, 2021.

\bibitem[Kalainathan(2019)]{kalainathan2019generative}
Diviyan Kalainathan.
\newblock \emph{Generative Neural Networks to infer Causal Mechanisms:
  algorithms and applications}.
\newblock PhD thesis, Universit{\'e} Paris Saclay (COmUE), 2019.

\bibitem[Kocaoglu et~al.(2017)Kocaoglu, Dimakis, Vishwanath, and
  Hassibi]{kocaoglu2017entropic}
Murat Kocaoglu, Alexandros~G Dimakis, Sriram Vishwanath, and Babak Hassibi.
\newblock Entropic causal inference.
\newblock In \emph{Thirty-First AAAI Conference on Artificial Intelligence},
  2017.

\bibitem[Kolmogorov(1968)]{kolmogorov1968three}
Andrei~Nikolaevic Kolmogorov.
\newblock Three approaches to the quantitative definition of information.
\newblock \emph{International journal of computer mathematics}, 2\penalty0
  (1-4):\penalty0 157--168, 1968.

\bibitem[Lopez-Paz et~al.(2015)Lopez-Paz, Muandet, Sch{\"o}lkopf, and
  Tolstikhin]{lopez2015towards}
David Lopez-Paz, Krikamol Muandet, Bernhard Sch{\"o}lkopf, and Iliya
  Tolstikhin.
\newblock Towards a learning theory of cause-effect inference.
\newblock In \emph{International Conference on Machine Learning}, pp.\
  1452--1461. PMLR, 2015.

\bibitem[Marx \& Vreeken(2017)Marx and Vreeken]{marx2017telling}
Alexander Marx and Jilles Vreeken.
\newblock Telling cause from effect using mdl-based local and global
  regression.
\newblock In \emph{2017 IEEE international conference on data mining (ICDM)},
  pp.\  307--316. IEEE, 2017.

\bibitem[Marx \& Vreeken(2018)Marx and Vreeken]{marx2018causal}
Alexander Marx and Jilles Vreeken.
\newblock Causal inference on multivariate and mixed-type data.
\newblock In \emph{Joint European Conference on Machine Learning and Knowledge
  Discovery in Databases}, pp.\  655--671. Springer, 2018.

\bibitem[Marx \& Vreeken(2019)Marx and Vreeken]{marx2019identifiability}
Alexander Marx and Jilles Vreeken.
\newblock Identifiability of cause and effect using regularized regression.
\newblock In \emph{Proceedings of the 25th ACM SIGKDD International Conference
  on Knowledge Discovery \& Data Mining}, pp.\  852--861, 2019.

\bibitem[Marx \& Vreeken(2021)Marx and Vreeken]{marx2021formally}
Alexander Marx and Jilles Vreeken.
\newblock Formally justifying mdl-based inference of cause and effect.
\newblock \emph{arXiv preprint arXiv:2105.01902}, 2021.

\bibitem[Mitrovic et~al.(2018)Mitrovic, Sejdinovic, and
  Teh]{mitrovic2018causal}
Jovana Mitrovic, Dino Sejdinovic, and Yee~Whye Teh.
\newblock Causal inference via kernel deviance measures.
\newblock \emph{Advances in neural information processing systems}, 31, 2018.

\bibitem[Mooij et~al.(2009)Mooij, Janzing, Peters, and
  Sch{\"o}lkopf]{mooij2009regression}
Joris Mooij, Dominik Janzing, Jonas Peters, and Bernhard Sch{\"o}lkopf.
\newblock Regression by dependence minimization and its application to causal
  inference in additive noise models.
\newblock In \emph{Proceedings of the 26th annual international conference on
  machine learning}, pp.\  745--752, 2009.

\bibitem[Mooij et~al.(2016{\natexlab{a}})Mooij, Peters, Janzing, Zscheischler,
  and Sch\"{o}lkopf]{cause_effect_moiij}
Joris~M. Mooij, Jonas Peters, Dominik Janzing, Jakob Zscheischler, and Bernhard
  Sch\"{o}lkopf.
\newblock {Distinguishing Cause from Effect Using Observational Data: Methods
  and Benchmarks}.
\newblock \emph{J. Mach. Learn. Res.}, 17\penalty0 (1):\penalty0 1103–1204,
  jan 2016{\natexlab{a}}.
\newblock ISSN 1532-4435.

\bibitem[Mooij et~al.(2016{\natexlab{b}})Mooij, Peters, Janzing, Zscheischler,
  and Sch{\"o}lkopf]{mooij2016distinguishing}
Joris~M Mooij, Jonas Peters, Dominik Janzing, Jakob Zscheischler, and Bernhard
  Sch{\"o}lkopf.
\newblock Distinguishing cause from effect using observational data: methods
  and benchmarks.
\newblock \emph{The Journal of Machine Learning Research}, 17\penalty0
  (1):\penalty0 1103--1204, 2016{\natexlab{b}}.

\bibitem[Nakkiran \& Bansal(2020)Nakkiran and
  Bansal]{nakkiran2020distributional}
Preetum Nakkiran and Yamini Bansal.
\newblock Distributional generalization: A new kind of generalization.
\newblock \emph{arXiv preprint arXiv:2009.08092}, 2020.

\bibitem[Park \& Boyd(2017)Park and Boyd]{park2017general}
Jaehyun Park and Stephen Boyd.
\newblock General heuristics for nonconvex quadratically constrained quadratic
  programming.
\newblock \emph{arXiv preprint arXiv:1703.07870}, 2017.

\bibitem[Pearl(2009)]{pearl2009causality}
Judea Pearl.
\newblock \emph{Causality}.
\newblock Cambridge university press, 2009.

\bibitem[Peters et~al.(2016)Peters, B{\"u}hlmann, and
  Meinshausen]{peters2016causal}
Jonas Peters, Peter B{\"u}hlmann, and Nicolai Meinshausen.
\newblock {Causal inference by using invariant prediction: identification and
  confidence intervals}.
\newblock \emph{Journal of the Royal Statistical Society: Series B (Statistical
  Methodology)}, 78\penalty0 (5):\penalty0 947--1012, 2016.

\bibitem[Peters et~al.(2017)Peters, Janzing, and
  Sch{\"o}lkopf]{peters2017elements}
Jonas Peters, Dominik Janzing, and Bernhard Sch{\"o}lkopf.
\newblock \emph{{Elements of causal inference: foundations and learning
  algorithms}}.
\newblock The MIT Press, 2017.

\bibitem[Rissanen(1978)]{rissanen1978modeling}
Jorma Rissanen.
\newblock Modeling by shortest data description.
\newblock \emph{Automatica}, 14\penalty0 (5):\penalty0 465--471, 1978.

\bibitem[Sch{\"o}lkopf et~al.(2012)Sch{\"o}lkopf, Janzing, Peters, Sgouritsa,
  Zhang, and Mooij]{scholkopf2012causal}
Bernhard Sch{\"o}lkopf, Dominik Janzing, Jonas Peters, Eleni Sgouritsa, Kun
  Zhang, and Joris Mooij.
\newblock On causal and anticausal learning.
\newblock \emph{arXiv preprint arXiv:1206.6471}, 2012.

\bibitem[Scott(1992)]{scott1992multivariate}
D.W. Scott.
\newblock \emph{Multivariate Density Estimation: Theory, Practice, and
  Visualization}.
\newblock A Wiley-interscience publication. Wiley, 1992.
\newblock ISBN 9780471547709.

\bibitem[Sgouritsa et~al.(2015)Sgouritsa, Janzing, Hennig, and
  Schölkopf]{sgouritsa2016inference}
Eleni Sgouritsa, Dominik Janzing, Philipp Hennig, and Bernhard Schölkopf.
\newblock {Inference of Cause and Effect with Unsupervised Inverse Regression}.
\newblock In Guy Lebanon and S.~V.~N. Vishwanathan (eds.), \emph{Proceedings of
  the Eighteenth International Conference on Artificial Intelligence and
  Statistics}, volume~38 of \emph{Proceedings of Machine Learning Research},
  pp.\  847--855, San Diego, California, USA, 09--12 May 2015. PMLR.
\newblock URL \url{https://proceedings.mlr.press/v38/sgouritsa15.html}.

\bibitem[Shimizu et~al.(2006)Shimizu, Hoyer, Hyv{\"a}rinen, Kerminen, and
  Jordan]{shimizu2006linear}
Shohei Shimizu, Patrik~O Hoyer, Aapo Hyv{\"a}rinen, Antti Kerminen, and Michael
  Jordan.
\newblock A linear non-gaussian acyclic model for causal discovery.
\newblock \emph{Journal of Machine Learning Research}, 7\penalty0 (10), 2006.

\bibitem[Spirtes et~al.(2000)Spirtes, Glymour, Scheines, and
  Heckerman]{spirtes2000causation}
Peter Spirtes, Clark~N Glymour, Richard Scheines, and David Heckerman.
\newblock \emph{Causation, prediction, and search}.
\newblock MIT press, 2000.

\bibitem[Sriperumbudur et~al.(2009)Sriperumbudur, Fukumizu, Gretton,
  Sch{\"o}lkopf, and Lanckriet]{sriperumbudur2009integral}
Bharath~K Sriperumbudur, Kenji Fukumizu, Arthur Gretton, Bernhard
  Sch{\"o}lkopf, and Gert~RG Lanckriet.
\newblock On integral probability metrics,$\backslash$phi-divergences and
  binary classification.
\newblock \emph{arXiv preprint arXiv:0901.2698}, 2009.

\bibitem[Stegle et~al.(2010)Stegle, Janzing, Zhang, Mooij, and
  Sch{\"o}lkopf]{stegle2010probabilistic}
Oliver Stegle, Dominik Janzing, Kun Zhang, Joris~M Mooij, and Bernhard
  Sch{\"o}lkopf.
\newblock Probabilistic latent variable models for distinguishing between cause
  and effect.
\newblock \emph{Advances in neural information processing systems}, 23, 2010.

\bibitem[Steininger et~al.(2021)Steininger, Kobs, Davidson, Krause, and
  Hotho]{steininger2021density}
Michael Steininger, Konstantin Kobs, Padraig Davidson, Anna Krause, and Andreas
  Hotho.
\newblock Density-based weighting for imbalanced regression.
\newblock \emph{Machine Learning}, 110\penalty0 (8):\penalty0 2187--2211, 2021.

\bibitem[Sun et~al.(2007)Sun, Janzing, and
  Sch{\"o}lkopf]{sun2007distinguishing}
Xiaohai Sun, Dominik Janzing, and Bernhard Sch{\"o}lkopf.
\newblock Distinguishing between cause and effect via kernel-based complexity
  measures for conditional distributions.
\newblock In \emph{15th European Symposium on Artificial Neural Networks (ESANN
  2007)}, pp.\  441--446. D-Side Publications, 2007.

\bibitem[Tagasovska et~al.(2018)Tagasovska, Chavez-Demoulin, and
  Vatter]{tagasovska2018distinguishing}
Natasa Tagasovska, Valérie Chavez-Demoulin, and Thibault Vatter.
\newblock Distinguishing {Cause} from {Effect} {Using} {Quantiles}: {Bivariate}
  {Quantile} {Causal} {Discovery}.
\newblock \emph{arXiv.org}, January 2018.
\newblock \doi{10.48550/arXiv.1801.10579}.
\newblock URL \url{https://arxiv.org/abs/1801.10579v4}.

\bibitem[Vowels et~al.(2021)Vowels, Camgoz, and Bowden]{vowels2021d}
Matthew~J Vowels, Necati~Cihan Camgoz, and Richard Bowden.
\newblock D'ya like dags? a survey on structure learning and causal discovery.
\newblock \emph{arXiv preprint arXiv:2103.02582}, 2021.

\bibitem[Wen et~al.(2018)Wen, Hassanpour, and Greiner]{wen2018weighted}
Junfeng Wen, Negar Hassanpour, and Russell Greiner.
\newblock Weighted gaussian process for estimating treatment effect.
\newblock In \emph{Proceedings of the 30th Annual Conference on Neural
  Information Processing Systems}, 2018.

\bibitem[Zhang \& Hyvarinen(2012)Zhang and Hyvarinen]{zhang2012identifiability}
Kun Zhang and Aapo Hyvarinen.
\newblock On the identifiability of the post-nonlinear causal model.
\newblock \emph{arXiv preprint arXiv:1205.2599}, 2012.

\end{thebibliography}
\bibliographystyle{iclr2023_conference}

\newpage
\begin{center}
	{\large Supplementary Material for the paper}
	
	{\LARGE Variation-based Cause Effect Identification}
	
\end{center}
\bigskip
\appendix
\section*{Appendix}
\section{Empirical Distribution}
\label{appendix:empirical-distribution}

The empirical probability density function (ePDF):
\begin{equation}
p_{x,N}(x) = \frac{1}{N}\sum_{n=1}^N \delta_{x_n}(x)
\end{equation}
is the derivative of the empirical cumulative distribution (eCDF) defined by
\begin{equation}
F_{x,N}({x}) = \frac{1}{N} \sum_{n=1}^{N} \mathds{1}_{{x}_n \,\leq\, {x}}
\end{equation}
where $\mathds{1}_{(\cdot)}$ is the indicator function and the inequality is to be understood entry-wise.
The eCDF $F_{x,N}(x)$ is the minimum variance unbiased estimator of the true CDF function $F_x(x)$ \citep{scott1992multivariate}.
The ePDF can also be viewed a limit case of kernel-density estimation.

The motivation behind such a modeling choice is that, we normally do not have the output of our \emph{unknown} system/data-generation-process to an arbitrary input $x$ (other than the samples pairs $\{x_n,y_n\}_{n=1}^N$.
Hence, in our search for a distinct marginal on e.g. $p_x$, we are limited to the convex set defined by the mixture distribution.
These are the stimuli for which we know the output of our unknown system treating it as a stochastic mapping.
This, in turn, allows us to treat the obtained weight vector as a sample weight on the joint distribution $p_{xy}$ and train models to approximate the conditionals $p_{x|y}$ and $p_{y|x}$ accordingly.

One downside is that the search space for a distinct marginal is limited to this convex set, which is itself sensitive to the sampling error.
A standard kernel density estimation can alleviate such a problem, but as mentioned, we assume no access to (nor information on) the underlying system allowing us to use this kde-based estimates on the output or joint spaces.

\section{Maximally Distinct Mixture}
\label{appendix:derivation}
In this section we detail the derivation of the \underline{s}emi\underline{d}efinite \underline{r}elaxation (SDR) approach to the optimization problem used in our method \cref{eq:original-formulation:objective}--\ref{eq:original-formulation:inequality}.

\subsection{From the Uniform Empirical}
\label{ch:mdm:uniform}
\paragraph{Problem 1}\emph{Given a set of samples $\mathcal{D}_{x} = \{{x}_n\}_{n=1}^N$ from a random variable $x\in\mathbb{X}$, find the weight vector $\bm{\alpha}$ that renders the mixture distribution $p_{x,N}^{\bm{\alpha}}$ maximally distinct from $p_{x,N}$ in some discrepancy measure $D(\cdot, \cdot)$.\label{problem1}}

With the kernel-based MMD measure $D\equiv \text{MMD}_{k_\mathbb{X}}$, Problem 1 can be formalized as
\begin{subequations}
\begin{align}
~~\underset{\bm{\alpha}}{\text{maximize}} ~~~~~ & \text{MMD}^2_{k_\mathbb{X}}(p_{x,N}^{\bm{\alpha}},\,p_{x,N}) 
\label{eq:original-formulation:objective_a}\\
\text{subject to} ~~~~~ & \bm{1}_N^\top \bm{\alpha} = 1 
\label{eq:original-formulation:equality_a}\\
& \bm{\alpha} \geqslant 0 \;\; \text{(entry-wise)
	\label{eq:original-formulation:inequality_a}}
\end{align}
\end{subequations}

where $\bm{1}_{N}$ refers to a vector of ones with dimensionality $N$. The quantity being optimized can be reformulated as follows:
\begin{subequations}
	\label{prob1_ref}
\begin{alignat}{2}
&\! \text{MMD}^2_{k_\mathbb{X}}(p_{x,N}^{\bm{\alpha}},\,p_{x,N})  &      =& \|p_{x,N}^{\bm{\alpha}}(x)-p_{x,N}(x)\|_{\mathcal{H}}^2\\
&                           &      =& \left\|\sum_{n=1}^{N}\alpha \delta_{\bm{x}_n} - \frac{1}{N} \sum_{n=1}^{N} \delta_{\bm{x}_n} \right\|_{\mathcal{H}}^2\\
&                           &      =& \sum_{n,n'=1}^{N}\alpha_n\alpha_{n'}\langle\delta_{\bm{x}_n},\delta_{\bm{x}_{n'}}\rangle 
	- \frac{2}{N}\sum_{n,n'=1}^{N}\alpha_n\langle\delta_{\bm{x}_n},\delta_{\bm{x}_{n'}}\rangle +\frac{1}{N^2}\sum_{n,n'=1}^{N}\langle\delta_{\bm{x}_n},\delta_{\bm{x}_{n'}}\rangle\\
&               &      =& \bm{\alpha}^\top \mathbf{K}_{xx} \bm{\alpha} 
- \frac{2}{N} \bm{\alpha}^\top \mathbf{K}_{xx} \bm{1}_N
+ \frac{1}{N^2} \bm{1}_N^\top \mathbf{K}_{xx} \bm{1}_N
\end{alignat}
\end{subequations}
where $\mathbf{K}_{xx}=[k(x_i,x_j)]_{i,j=1}^N$ is the Gram matrix of the kernel function $k_\mathbb{X}: \mathbb{X}\times\mathbb{X}\to\mathbb{R}^+$ on the sample set $\mathcal{D}_{x}$. 
with which the optimization problem becomes:  
\begin{subequations}
\begin{alignat}{2}
~~\underset{\bm{\alpha}}{\text{maximize}} ~~~~~ &
\bm{\alpha}^\top \mathbf{K}_{xx} \bm{\alpha} 
- \frac{2}{N} \bm{\alpha}^\top \mathbf{K}_{xx} \bm{1}_N
+ \frac{1}{N^2} \bm{1}_N^\top \mathbf{K}_{xx} \bm{1}_N \label{19a} \\
\text{subject to} ~~~~~ & \bm{1}_N^\top \bm{\alpha} = 1 \\
& \bm{\alpha} \geqslant 0 \;\; \text{(entry-wise)}
\end{alignat}
\end{subequations}

The optimization problem is not a convex optimization problem since it is a \emph{maximization} of a convex function.
Noting that the closed-form estimator of the squared MMD has a quadratic form in the optimization variable $\bm{\alpha}$, \citet{park2017general} address this problem in a two-step procedure referred to as \underline{s}emi\underline{d}efinite \underline{r}elaxation (SDR). 
They first \emph{lift} the problem to a higher dimensional space by defining $\mathbf{A}=\bm{\alpha}\bm{\alpha}^\top$ in which the objective function becomes linear, then apply a convex \emph{relaxation} to the intractable constraints.
Without affecting the solution to the problem and using the properties of the \textbf{trace} of a matrix, each term of the objective \cref{19a} can be reformulated as:
\begin{subequations}
\begin{alignat}{2}
\bm{\alpha}^\top \mathbf{K}_{xx} \bm{\alpha} 
	&= \text{\textbf{trace}}(\bm{\alpha}^\top \mathbf{K}_{xx} \bm{\alpha})\\
	&= \text{\textbf{trace}}(\bm{\alpha} \bm{\alpha}^\top \mathbf{K}_{xx} )\\
	&= \text{\textbf{trace}}(\mathbf{A} \mathbf{K}_{xx} )\\
    &= \mathbf{A} \bullet \mathbf{K}_{xx}
\end{alignat}
\end{subequations}
and similarly for the second term:
\begin{subequations}
\begin{alignat}{2}
\bm{\alpha}^\top \mathbf{K}_{xx} \bm{1}_N  
	&= \text{\textbf{trace}}(\bm{\alpha}^\top \mathbf{K}_{xx} \bm{1}_N) \\
    &= \text{\textbf{trace}}(\bm{\alpha} \bm{\alpha}^\top \mathbf{K}_{xx} \bm{1}_N\bm{1}_N^\top) \\
    &= \mathbf{A} \bullet \mathbf{K}_{xx} \bm{1}_N\bm{1}_N^\top
\end{alignat}
\end{subequations}
where $\bullet$ denotes the dot-product in matrix space defined as $\mathbf{A}\bullet\mathbf{K}_{xx} = \text{\textbf{trace}}(\mathbf{A}\mathbf{K}_{xx})$.
They then extract all convex constraints from the condition $\mathbf{A}=\bm{\alpha}\bm{\alpha}^\top = [a_{ij}]_{i,j=1}^{N,N} $
The first is the entry-wise non-negativity $a_{ij} = \alpha_i \alpha_j \geqslant 0$ due to the entry-wise non-negativity of $\bm{\alpha}\in[0,1]^N$.
The second is the consequence of the normalized vector $\bm{1}_N^\top\bm{\alpha}=1$ which can expressed in $\mathbf{A}$ as $\bm{1}_N^\top\mathbf{A}\bm{1}=\bm{1}_N^\top\bm{\alpha}(\bm{1}_N^\top\bm{\alpha})^\top = 1$.
The last is the similarity of $\mathbf{A}=\mathbf{A}^\top$ by definition.
Finally, the equality condition above is relaxed to $\mathbf{A}\succeq \bm{\alpha}\bm{\alpha}^\top$ and written in its Schur-complement form.

As a result, the following formulation is a relaxation of \ref{eq:original-formulation:objective_a}--\ref{eq:original-formulation:inequality_a} which is a \underline{q}uadratically \underline{c}onstraint \underline{q}uadratic \underline{p}rogram (QCQP):
\begin{align}
~~\underset{\mathbf{A}}{\text{maximize}} ~~~~~ &  \mathbf{A} \bullet \left(\mathbf{K}_{xx}-\frac{2}{N}\mathbf{K}_{xx}\bm{1}_N\bm{1}_N^\top \right) + \frac{1}{N^2} \bm{1}_N^\top\mathbf{K}_{xx}\bm{1}_N 
\label{eq:sdr-formulatoin:objective_a}   \\
\text{subject to} ~~~~~ &  \begin{bmatrix}
\mathbf{A} 					& \mathbf{A}\bm{1}_N \\
\bm{1}_N^\top\mathbf{A}		& 1 \\
\end{bmatrix} \;\succeq \;0   \quad \text{(positive semidefiniteness)}
\label{eq:sdr-formulation:inequality:psd_a}  \\
& \mathbf{A} \geqslant 0 ~\qquad\qquad\qquad \text{(entry-wise)}
\label{eq:sdr-formulation:inequality:entry_a} \\
& \bm{1}_N^\top\mathbf{A}\bm{1}_N = 1 
\label{eq:sdr-formulation:equality:normalization_a} \\
& \mathbf{A} = \mathbf{A}^\top 
\label{eq:sdr-formulation:equality:symmetry_a}      									   
\end{align}
this problem has a convex object (linear) with convex constraints which can be solved using existing packages such as \texttt{cvxpy} \cite{diamond2016cvxpy}.

\paragraph{Problem 2:}\emph{Given two sets of samples $\{\bm{x}_n\}_{n=1}^N$ and $\{\bm{\tilde{x}}_m\}_{m=1}^M$ from the two distributions $p_{x,N}$ and $p_{\tilde{x},M}$, respectively, with the corresponding random variables $x,\tilde{x}\in\mathbb{X}$ find the weight vector $\tilde{\bm{\alpha}}\in[0,1]^M$ that renders the mixture distribution $p_{\tilde{x},M}^{\tilde{\bm{\alpha}}}$ maximally distinct from $p_{x,N}$} w.r.t the discrepancy measure $\text{MMD}_{k_\mathbb{X}}$.

This problem can be formalized as
\begin{subequations}
	\begin{align}
		~~\underset{\bm{\alpha}}{\text{maximize}} ~~~~~ & \text{MMD}^2_{k_\mathbb{X}}(p_{\tilde{x},M}^{\tilde{\bm{\alpha}}},\,p_{x,N}) 
		\label{eq:original-formulation:objective_a2}\\
		\text{subject to} ~~~~~ & \bm{1}_M^\top \tilde{\bm{\alpha}} = 1 
		\label{eq:original-formulation:equality_a2}\\
		& \tilde{\bm{\alpha}} \geqslant 0 \;\; \text{(entry-wise)
			\label{eq:original-formulation:inequality_a2}}
	\end{align}
\end{subequations}

Same as in \ref{prob1_ref} the objective can be reformulated as follows:
\begin{subequations}
\begin{align}
\text{MMD}^2_{k_\mathbb{X}}(p_{\tilde{x},M}^{\tilde{\bm{\alpha}}},\,p_{x,N})    & =  \left\|p_{\tilde{x},M}^{\tilde{\bm{\alpha}}}(\tilde{x})-p_{x,N}(x)\right\|^2_\mathcal{H}\\
& = \tilde{\bm{\alpha}}^\top \mathbf{K}_{\tilde{x}\tilde{x}} \tilde{\bm{\alpha}} - \frac{2}{N} \tilde{\bm{\alpha}}^\top \mathbf{K}_{\tilde{x}x} \bm{1}_N + \frac{1}{N^2} \bm{1}_N^\top \mathbf{K}_{xx} \bm{1}_N
\end{align}
\end{subequations}
Similar to Problem 1, the objective terms can be rewritten as:
\begin{subequations}
	\begin{align}
		\tilde{\bm{\alpha}}^\top \mathbf{K}_{\tilde{x}\tilde{x}} \tilde{\bm{\alpha}} 
		&= \tilde{\mathbf{A}} \bullet \mathbf{K}_{\tilde{x}\tilde{x}}
	\end{align}
\end{subequations}
and similarly for the second term:
\begin{subequations}
	\begin{align}
		\tilde{\bm{\alpha}}^\top \mathbf{K}_{\tilde{x}x} \bm{1}_N  
		&= \tilde{\mathbf{A}} \bullet \mathbf{K}_{\tilde{x}x} \bm{1}_N\bm{1}_N^\top
	\end{align}
\end{subequations}
The constraints can be modified as in Problem 1. Hence, a relaxation of \ref{eq:original-formulation:objective_a2}--\ref{eq:original-formulation:inequality_a2} is formulated as:
\begin{align}
	~~\underset{\tilde{\mathbf{A}}}{\text{maximize}} ~~~~~ &  \tilde{\mathbf{A}} \bullet \left(\mathbf{K}_{\tilde{x}\tilde{x}}-\frac{2}{N}\mathbf{K}_{\tilde{x}x}\bm{1}_N\bm{1}_N^\top \right) + \frac{1}{N^2} \bm{1}_N^\top\mathbf{K}_{xx}\bm{1}_N 
	\label{eq:sdr-formulatoin:objective_a2}   \\
	\text{subject to} ~~~~~ &  \begin{bmatrix}
		\tilde{\mathbf{A}} 					& \tilde{\mathbf{A}}\bm{1}_M \\
		\bm{1}_M^\top\tilde{\mathbf{A}}		& 1 \\
	\end{bmatrix} \;\succeq \;0   \quad \text{(positive semidefiniteness)}
	\label{eq:sdr-formulation:inequality:psd_a2}  \\
	& \tilde{\mathbf{A}} \geqslant 0 ~\qquad\qquad\qquad \text{(entry-wise)}
	\label{eq:sdr-formulation:inequality:entry_a2} \\
	& \bm{1}_M^\top\tilde{\mathbf{A}}\bm{1}_M = 1 
	\label{eq:sdr-formulation:equality:normalization_a2} \\
	& \tilde{\mathbf{A}} = \tilde{\mathbf{A}}^\top 
	\label{eq:sdr-formulation:equality:symmetry_a2}      									   
\end{align}
which is a QCQP on the $M^2$ optimization variables in $\tilde{\mathbf{A}}=[\tilde{a}_{ij}]_{i,j=1}^{M,M}$.

\section{Experimental Setup and Further Analysis}
\label{appendix:experiment-setup}

In this section we detail the experimental setup that was used in estimating the results presented in \cref{fig:rel_acc}.
We first standardize the dataset using the \texttt{RobustScaler}from the \texttt{sklearn} library [B1]. 
As a second step we extract randomly $M$ samples to use further in the optimization problem from \ref{sub:artificial-setups}. 
The next steps are then to be followed as stated in the Algorithm \ref{alg:VCEI} where the hyperparameters were defined as follows:
\begin{enumerate}
	\item We use a squared exponential kernel (SEK), with its maximum likelihood estimate of its lengthscale parameter using a KDE on a 5-fold cross validation scheme. 
	\item We use the Exact-GP as our predictive model class $\mathcal{M}$ (SEK as a kernel).
	\item We use $b_\alpha = 0.2$
	\item We use the mean value for the predition of the GP model
	\item All experiments took place on an 8-core processor from a single PC (without GPU compute power).
\end{enumerate}

Note that in case of a large dataset (such as the pair-07 in T\"ubingen benchmark) we extract a subset that represents the distribution of the original set, referred to as a coreset $\mathcal{D_\mathbf{C}}$ which is estimated as follows. From a KDE estimate [B2] on either of the marginals (on $x$ and $y$), include the $k$ \emph{rare} samples of with probability lower than 0.05 in either of the marginal KDEs.
This is then further complemented with $M-k$ samples drawn randomly.
This last step (the random draw of $M-k$ samples) is repeated a number of times, and the case with the minimal MMD to the original set is selected.
In case of a small dataset, the coreset is automatically identical to the main set.

\end{document}